\documentclass[11pt,a4paper]{article}
\usepackage[hyperref]{naaclhlt2018}
\usepackage{times}
\usepackage{latexsym}
\usepackage{graphicx}  
\usepackage{url}
\usepackage{lipsum}
\usepackage{breqn}
\usepackage{subfigure}
\usepackage{paralist}
\usepackage{enumitem}
\usepackage{arydshln}
\usepackage{lingmacros}
\usepackage{booktabs}

\aclfinalcopy 

\begin{document}
\title{Resolving Event Coreference with Supervised Representation Learning and Clustering-Oriented Regularization}

\author{%
        Kian Kenyon-Dean \\
        School of Computer Science\\
	    McGill University\\
	    {\tt kian.kenyon-dean}\\ {\tt@mail.mcgill.ca}
    \And
        Jackie Chi Kit Cheung\\
        School of Computer Science\\
	    McGill University\\
	    {\tt jcheung@cs.mcgill.ca}
    \And
        Doina Precup\\
        School of Computer Science\\
	    McGill University\\
	    {\tt dprecup@cs.mcgill.ca}}

\maketitle
\begin{abstract}
We present an approach to event coreference resolution by developing a general framework for clustering that uses supervised representation learning. We propose a neural network architecture with novel Clustering-Oriented Regularization (CORE) terms in the objective function. These terms encourage the model to create embeddings of event mentions that are amenable to clustering. We then use agglomerative clustering on these embeddings to build event coreference chains. For both within- and cross-document coreference on the ECB+ corpus, our model obtains better results than models that require significantly more pre-annotated information. This work provides insight and motivating results for a new general approach to solving coreference and clustering problems with representation learning.
\end{abstract}

\noindent 

\section{Introduction}
Event coreference resolution is the task of determining which \textit{event mentions} expressed in language refer to the same real-world event instances. The ability to resolve event coreference has improved the quality of downstream tasks such as automatic text summarization \cite{vanderwende2004event}, questioning-answering \cite{berant2014modeling}, headline generation \cite{sun2015event}, and text-mining in the medical domain \cite{ferracane2016leveraging}. 

Event mentions are comprised of an action component (or, head) and surrounding arguments. Consider the following passages, drawn from two different documents; the heads of the event mentions are in boldface and the subscripts indicate mention IDs:

\enumsentence{The president's \underline{\textbf{speech}}$_{m1}$ \textbf{shocked}$_{m2}$ the audience. He \underline{\textbf{announced}}$_{m3}$ several new controversial policies.}   
\enumsentence{The policies \underline{\textbf{proposed}}$_{m4}$ by the president will not \textbf{surprise}$_{m5}$ those who \textbf{followed}$_{m6}$ his \textbf{campaign}$_{m7}$.}

In this example, $m1$, $m3$, and $m4$ form a chain of coreferent event mentions (underlined), because they refer to the same real-world event in which the president gave a speech. The other four are singletons, meaning that they all refer to separate events and do not corefer with any other mention.
 
This work investigates how to learn useful representations of event mentions. Event mentions are complex objects, and both the event mention heads and the surrounding arguments are important for the event coreference resolution task. In our example above, the head words of mentions $m2$, \textit{shocked}, and $m5$, \textit{surprise}, are lexically similar, but the event mentions do not corefer. This task therefore necessitates a model that can capture the distributional relationships between event mentions and their surrounding contexts.

We hypothesize that prior knowledge about the task itself can be usefully encoded into the representation learning objective. For our task, this prior means that the embeddings of corefential event mentions should have similar embeddings to each other (a ``natural clustering'', using the terminology of \newcite{bengio2013representation}). With this prior, our model creates embeddings of event mentions that are directly conducive for the clustering task of building event coreference chains. This is contrary to the indirect methods of previous work that rely on pairwise decision making followed by a separate model that aggregates the sometimes inconsistent decisions into clusters (Section~\ref{sec:relwork}).

We demonstrate these points by proposing a method that learns to embed event mentions into a space that is tuned specifically for clustering. The representation learner is trained to predict which event cluster the event mention belongs to, using an hourglass-shaped neural network. We propose a mechanism to modulate this training by introducing \textit{Clustering-Oriented Regularization} (CORE) terms into the objective function of the learner; these terms impel the model to produce similar embeddings for coreferential event mentions, and dissimilar embeddings otherwise.

Our model obtains strong results on within- and cross-document event coreference resolution, matching or outperforming the system of \citeauthor{cybulska2015translating} \shortcite{cybulska2015translating} on the ECB+ corpus on all six evaluation measures. We achieve these gains despite the fact that our model requires significantly less pre-annotated or pre-detected information in terms of the internal event structure. Our model's improvements upon the baselines show that our supervised representation learning framework creates new embeddings that capture the abstract distributional relations between samples and their clusters, suggesting that our framework can be generalized to other clustering tasks\footnote{All code used in this paper can be found here:\\ \url{https://github.com/kiankd/events}}.

\section{Related Work} \label{sec:relwork}
The recent work on event coreference can be categorized according to the assumed level of event representation. In the predicate-argument alignment paradigm \cite{roth2012aligning,wolfe2013parma}, links are simply drawn between predicates in different documents. This work only considers cross-document event coreference \cite{wolfe2013parma,wolfe2015predicate}, and no within-document coreference. At the other extreme, the ACE and ERE datasets annotate rich internal event structure, with specific taxonomies that describe the annotated events and their types \cite{ldc2005,ldc2016}. In these datasets, only within-document coreference is annotated.

The creators of the ECB \cite{bejan2008linguistic} and ECB+ \cite{cybulska2014using}, annotate events according to a level of abstraction between that of the predicate-argument approach and the ACE approach, being most similar to the TimeML paradigm \cite{pustejovsky2003timeml}. In these datasets, both within-document and cross-document coreference relations are annotated. We use the ECB+ corpus in our experiments because it solves the lack of lexical diversity found within the ECB by adding 502 new annotated documents, providing a total of 982 documents. 

Previous work on model design for event coreference has focused on clustering over a linguistically rich set of features. Most models require a pairwise-prediction based supervised learning step which predicts whether or not a pair of event mentions is coreferential \cite{bagga1999cross,chen2009pairwise,cybulska2015translating}. Other work focuses on the clustering step itself, aggregating local pairwise decisions into clusters, for example by graph partitioning \cite{chen2009graph}. There has also been work using non-parametric Bayesian clustering techniques \cite{bejan2014unsupervised,yang2015hierarchical}, as well as other probabilistic models \cite{lu2017learning}. Some recent work uses intuitions combining representation learning with clustering, but does not augment the loss function for the purpose of building clusterable representations \cite{krause2016event,choubey2017event}.



\section{Event Coreference Resolution Model} \label{sec:evco}
We formulate the task of event coreference resolution as creating clusters of event mentions which refer to the same event. For the purposes of this work, we define an event mention to be a set of tokens that correspond to the \textit{action} of some event. Consider the sentence below (borrowed from \citeauthor{cybulska2014using} \shortcite{cybulska2014using}): 

\enumsentence{On Monday Lindsay Lohan \textbf{checked into} rehab in Malibu, California after a car \textbf{crash}.}\label{ex:lindsay}

Our model would take, as input, feature vectors (see Section \ref{sec:features}) extracted from the two event mentions (in bold) independently. In this paper, we use the gold-standard event mentions provided by the dataset, and leave mention detection to other work.

\subsection{Model Overview}
Our approach to resolving event coreference consists of the following steps:
\begin{enumerate}[leftmargin=*]
    \item Train a supervised neural network model which learns event mention embeddings by predicting the event cluster in the training set to which the mention belongs (Figure~\ref{fig:train_model}).
    \item At test time, use the previously trained model's embedding layer to derive representations of unseen event mentions. Then, perform agglomerative clustering with these embeddings to create event coreference chains (Figure~\ref{fig:test_model}).
\end{enumerate}

\subsection{Supervised Representation Learning}
\label{sec:learner}
\begin{figure*}
    \centering    
    \begin{minipage}[b]{.49\textwidth}
    \includegraphics[width=0.9\columnwidth]{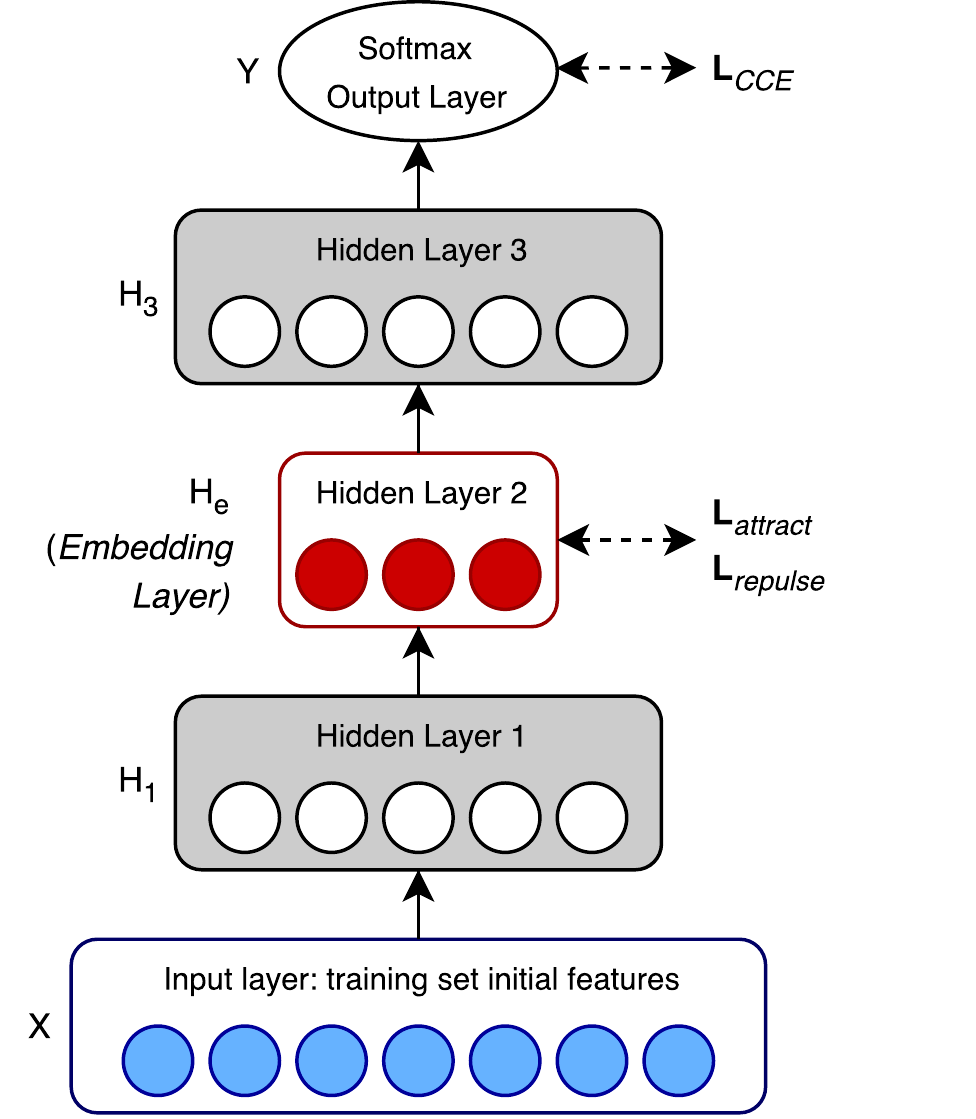}
    \caption{Our supervised representation learning model during the training step. Dashed arrows indicate contributions to the loss function.}\label{fig:train_model}
    \end{minipage}
    \hfill
    \begin{minipage}[b]{.49\textwidth}
    \includegraphics[width=0.9\columnwidth]{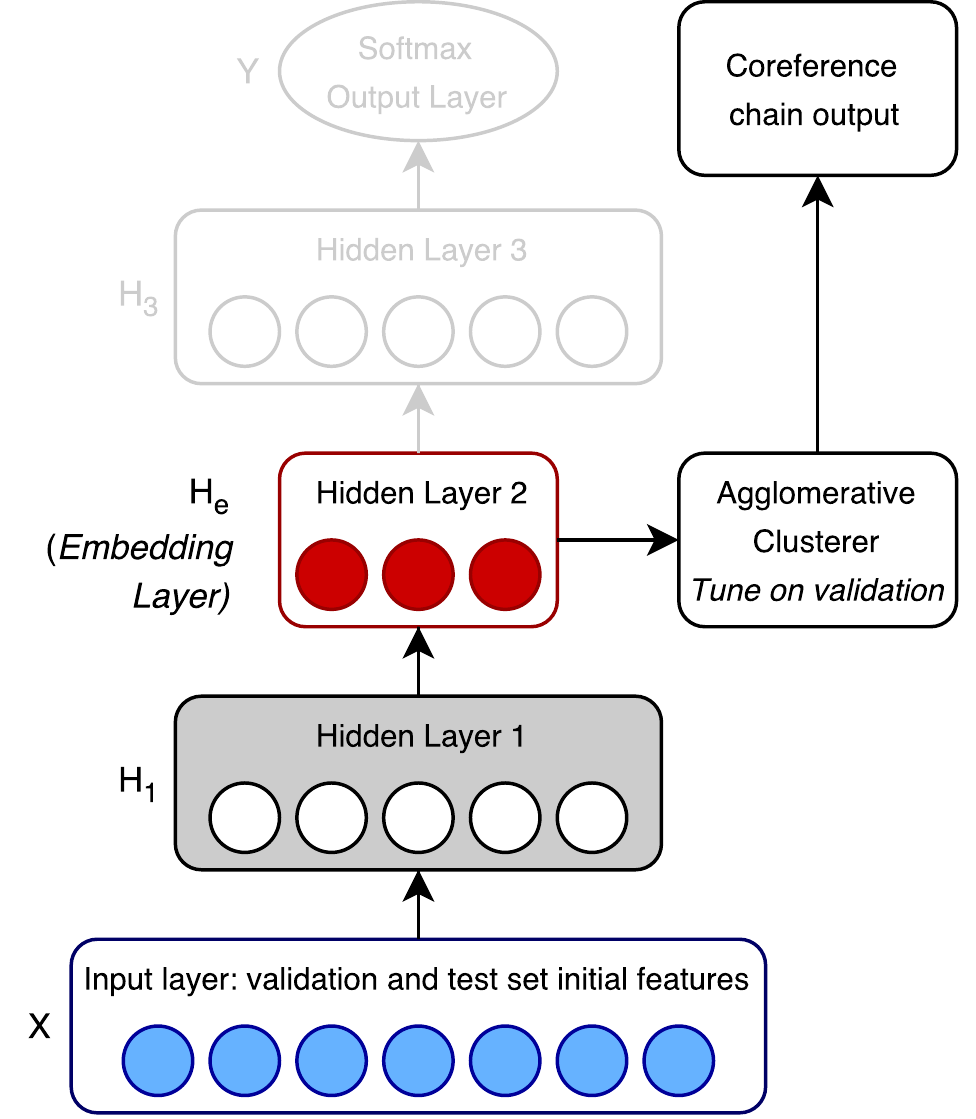}
    \caption{Our trained model at inference time, used for validation tuning and final testing. Note that $H_3$ and $Y$ are not used in this step.}\label{fig:test_model}
    \end{minipage}
\end{figure*}

We propose a representation learning framework based on training a multi-layer artificial neural network, with one layer H$_e$ chosen to be the embedding layer. In the training set, there are a certain number of event mentions, each of which belongs to some gold standard cluster, making $C$ total non-singleton clusters in the training set. The network is trained as if it were encountering a $C$+1-class classification problem, where the class of an event mention corresponds to a single output node, and all singleton mentions belong to class $C$+1\footnote{If each singleton mention (i.e., a mention that does not corefer with anything else) had its own class then the model would be confronted with a classification problem with thousands of classes, many of which would only have one sample; this is much too ill-posed, so we merge all singletons together during the training step.}. 

When using this model to cluster a new set of mentions, the final layer's output will not be directly informative since the output node structure corresponds to the clusters within the training set. However, we hypothesize that the trained model will have learned to capture the abstract distributional relationships between event mentions and clusters in the intermediate layer H$_e$. We thus use the activations in H$_e$ as the embedding of an event mention for the clustering step (see Figure~\ref{fig:test_model}). A similar hourglass-like neural architecture design has been successful in automatic speech recognition \cite{grezl2007probabilistic,gehring2013extracting}, but has not to our knowledge been used to pre-train embeddings for clustering.


\subsection{Categorical-Cross-Entropy (CCE)} Using CCE as the loss function trains the model to correctly predict a training set mention's corresponding cluster. With model prediction $y_{ic}$ as the probability that sample $i$ belongs to class $c$, and indicator variable $t_{ic} = 1$ if sample $i$ belongs to class $c$ (else $t_{ic} = 0$), we have the mean categorical-cross entropy loss over a randomly sampled training input batch $\mathrm{X}$:
\begin{dmath} \label{eq:cce}
 \textbf{L}_{CCE} = - \frac{1}{|\mathrm{X}|} \sum_{i=1}^{|\mathrm{X}|} \sum_{c=1}^{C+1} t_{ic} \log(y_{ic})
\end{dmath}

\subsection{Clustering-Oriented Regularization (CORE)}
With CCE, the model may overfit towards accurate prediction performance for those particular clusters found in the training set without learning an embedding that captures the nature of events in general. This therefore motivates introducing regularization terms based on the intuition that embeddings of mentions belonging to the same cluster should be similar, and that embeddings of mentions belonging to different clusters should be dissimilar. Accordingly, we define dissimilarity between two vector embeddings ($\vec{e_1},\vec{e_2}$) according to the cosine-distance function $\textbf{d}$:
%
\begin{equation} \label{eq:cosd}
    \textbf{d}(\vec{e_1}, \vec{e_2}) = \frac{1}{2} \Big(1 - \frac{\vec{e_1} \cdot \vec{e_2}}{||\vec{e_1}|| ||\vec{e_2}||}\Big)
\end{equation}

Given input batch $\mathrm{X}$, we create two sets $\mathcal{S}$ and $\mathcal{D}$, where $\mathcal{S}$ is the set of all pairs $(a,b)$ of mentions in $\mathrm{X}$ that belong to the same cluster, and $\mathcal{D}$ is the set of all pairs $(c,d)$ in $\mathrm{X}$ that belong to different clusters. Note that all vector embeddings $\vec{e_i} =$ H$_{e}(i)$; i.e., they are obtained by feeding the event mention $i$'s features through to embedding layer H$_{e}$. We now define the following \textit{Attractive} and \textit{Repulsive} CORE terms.

\subsubsection{Attractive Regularization} The first desirable property for the embeddings is that mentions that belong to the same cluster should have low cosine distance between each others' embeddings, since the agglomerative clustering algorithm uses cosine distance to make coreference decisions.

Formally, for all pairs of mentions $a$ and $b$ that belong to the same cluster, we would like to minimize the distance between their embeddings $\vec{e_a}$ and $\vec{e_b}$. We call this ``attractive'' regularization because we want to attract embeddings closer to each other by minimizing their distance $\textbf{d}(\vec{e_a}, \vec{e_b})$ so that they will be as similar as possible. 
\begin{dmath} \label{eq:attract}
 \textbf{L}_{\textit{attract}} = \frac{1}{|\mathcal{S}|}\sum_{(a,b) \in \mathcal{S}} \textbf{d}(\vec{e_a}, \vec{e_b})
\end{dmath}

\subsubsection{Repulsive Regularization} The second desirable property is that the embeddings corresponding to mentions that belong to different clusters should have high cosine distance between each other. Thus, for all pairs of mentions $c$ and $d$ that belong to different clusters, the goal is to maximize their distance $\textbf{d}(\vec{e_c}, \vec{e_d})$. This is ``repulsive'' because we train the model to push away the embeddings from each other to be as distant as possible.
\begin{dmath} \label{eq:repulse}
 \textbf{L}_{\textit{repulse}} = 1 - \frac{1}{|\mathcal{D}|}\sum_{(c,d) \in \mathcal{D}} \textbf{d}(\vec{e_c}, \vec{e_d})
\end{dmath}

\subsection{Loss Function} Equation \ref{eq:reg} below shows the final loss function\footnote{Note that, while we present Equations~\ref{eq:attract} and \ref{eq:repulse} as summations over pairs from the input batch, the computation is actually reasonable when written in terms of matrix multiplications. The most expensive operation multiplying the embedded batch of input samples times its transpose.}. The attractive and repulsive terms are weighted by hyperparameter constants $\lambda_1$ and $\lambda_2$ respectively:
\begin{dmath} \label{eq:reg}
    \textbf{L} = \textbf{L}_{\textit{CCE}} + \lambda_{1} \textbf{L}_{\textit{attract}} + \lambda_{2} \textbf{L}_{\textit{repulse}}
\end{dmath}


By adding these regularization terms to the loss function, we hypothesize that the new embeddings of test set mentions (obtained by feeding-forward their features into the trained model) will exemplify the desired properties represented by the loss function, thus assisting the agglomerative clustering task in producing correct coreference-chains.

\subsection{Agglomerative Clustering} 
Agglomerative clustering is a non-parametric ``bottom-up'' approach to hierarchical clustering, in which each sample starts as its own cluster, and at each step, the two most similar clusters are merged, where similarity between two clusters is measured according to some similarity metric. After each merge, clustering similarities are recomputed according to a preset criterion (e.g., single- or complete-linkage). In our models, clustering proceeds until a pre-determined similarity threshold, $\tau$, is reached. We tuned $\tau$ on the validation set, doing grid search for $\tau \in [0, 1]$ to maximize B$^3$ accuracy\footnote{We optimize with B$^3$ F1-score because the other measures are either too expensive to compute (CEAF-M, CEAF-E, BLANC), or are less discriminative (MUC).}. Preliminary experimentation led us to use cosine-similarity (see cosine distance in Equation~\ref{eq:cosd}) to measure vector similarity, and single-linkage for clustering decisions.

We experimented with two initialization schemes for agglomerative clustering. In the first scheme, each event mention is initialized as its own cluster, as is standard. In the second, we initialized clusters using the lemma-$\delta$ baseline defined by \citeauthor{upadhyay2016revisiting} \shortcite{upadhyay2016revisiting}. This baseline merges all event mentions with the same head lemma that are in documents with document-level similarity that is higher than a threshold $\delta$. \citeauthor{upadhyay2016revisiting} showed that it is a strong indicator of event coreference, so we experimented with initializing our clustering algorithm in this way. We call this model variant \textsc{CORE+CCE+Lemma}, and describe the parameter tuning procedures in more detail in Section \ref{sec:exp}.

\section{Feature Extraction} \label{sec:features}

\begin{table}[t]
    \begin{center}
        \begin{tabular}{|l|l|}
        \hline
        1.action      & \textit{checked into, crash}\\ \hline
        2.time        & \textit{On Monday}\\ \hline
        3.location    & \textit{rehab in Malibu, California} \\ \hline
        4.participant & \textit{Lindsay Lohan} (human) \\ 
                      & \textit{car} (non-human) \\ \hline
        \end{tabular}
    \end{center}
    \caption{An event template of the sentence in Example~\ref{ex:lindsay}, borrowed from \citeauthor{cybulska2014using} \shortcite{cybulska2014using,cybulska2015translating}. Our model only requires as input the \textit{action}, not the \textit{time}, \textit{location}, nor \textit{participant} arguments.}\label{tab:eventTemplate}
\end{table}

\noindent We extract features that do not require the pre-processing step of event-template construction to represent the context (unlike \citeauthor{cybulska2015translating} \shortcite{cybulska2015translating}, see Table~\ref{tab:eventTemplate}); instead, we represent the surrounding context by using the tokens in the general vicinity of the event's action. We thus only require the event's action -- which is what we define as an \textit{event mention} -- to be previously detected, not all of its arguments. We motivate this by arguing that it would be preferable to build high quality coreference chains without event template features since since extracting event templates can be a difficult process, with the possibility of errors cascading into the event coreference step. 

\subsection{Contextual} Inspired by the approach of  \citeauthor{clark2016improving} \shortcite{clark2016improving} in the entity coreference task, we extract, for the token sets below, (i) the token's \textit{word2vec} word embedding \cite{mikolov2013distributed} (or average if there are multiple); and, (ii) the one-hot count vector of the token's lemma\footnote{This is a $500$-dimensional vector where the first $499$ entries correspond to the $499$ most frequently occurring lemmas in the training set, and the $500^{th}$ entry indicates if the lemma is not in that set of most frequently occurring lemmas.} (or sum if there are multiple), for each event mention, $em$:


\begin{itemize}[nosep]
    \item the first token of $em$;
    \item the last token of $em$;
    \item all tokens in the $em$;
    \item each of the two tokens preceding $em$;
    \item each of the two tokens following $em$;
    \item all of the five tokens preceding $em$;
    \item all of the five tokens following $em$;
    \item all of the tokens in $em$'s sentence.
\end{itemize}

\subsection{Document} It is necessary to include features that characterize the mention's document, hoping that the model learns a latent understanding of relations between documents.  We extract features from the event mention's document by building lemma-based TF-IDF vector representations of the document. We use log normalization of the raw term frequency of token lemma $t$ in document $d$, $f_{t,d}$, where $\textit{TF}_t = 1+\log(f_{t,d})$. For the IDF term we use smoothed inverse document frequency, with $N$ as the number of documents and $n_t$ as the number of documents that contain the lemma, we have $\textit{IDF}_t = \log(1+ \frac{N}{n_t})$. By performing a component-wise multiplication of the $\textit{IDF}$ vector with each row in term-frequency matrix $\textit{TF}$, we create TF-IDF vectors of each document in the training and test sets (with length corresponding to the number of unique lemmas in the training set). We compress these vectors to $100$ dimensions with principal component analysis fitted onto the train set document vectors, which is used to transform the validation and test set document vectors.

\subsection{Comparative} We include comparative features to relate a mention to the other mentions in its document and to the mentions in the set of documents the model would be requested to extract event coreference chains from. This is motivated by the fact that coreference decisions must be informed by the relationship mentions have with each other. Firstly, we encode the position of the mention in its document with specific binary features indicating if it is first or last; for example, if there were five mentions and it were the third, this feature would correspond to the vector $[0, \frac{3}{5}, 0]$. 

Next, we define two sets of mentions we would like to compare with: the first contains all mentions in the same document as the current mention $em$, and the second contains all mentions in the data we are asked to cluster. For each of these sets, we compute: the average word overlap and average lemma overlap (measured by harmonic similarity) between $em$ and each of the other mentions in the set. We thus add two feature vector entries for each of the sets: the average word overlap between $em$ and the other mentions in the set, and the average lemma overlap between $em$ and the other mentions in the set.

\section{Experimental Design} \label{sec:exp}
We run our experiments on the ECB+ corpus, the largest corpus that contains both within- and cross-document event coreference annotations. We followed the train/test split of \citeauthor{cybulska2015translating} \shortcite{cybulska2015translating}, using topics $1$-$35$ as the train set and $36$-$45$ as the test set. During training, we split off a validation set\footnote{Topics 2, 5, 12, 18, 21, 23, 34, 35 (randomly chosen).} for hyperparameter tuning. 



Following \citeauthor{cybulska2015translating}, we used the portion of the corpus that has been manually reviewed and checked for correctness. Some previous work \cite{yang2015hierarchical,upadhyay2016revisiting,choubey2017event} do not appear to have followed this guideline from the corpus creators, as they report different corpus statistics compared to those reported by \citeauthor{cybulska2015translating}. As a result, those papers may report results on a data set with known annotation errors.

\subsection{Evaluation Measures}
Since there is no consensus in the coreference resolution literature on the best evaluation measure, we present results obtained according to six different measures, as is common in previous work. We use the scorer presented by \citeauthor{pradhan2014scoring} \shortcite{pradhan2014scoring}. In this task, the term ``coreference chain'' is synonymous with ``cluster''.
\vspace{0.5em}

\noindent
\textbf{MUC} \cite{vilain1995model}. Link-level measure which counts the minimum number of link changes required to obtain the correct clustering from the predictions; it does not account for correctly predicted singletons.
\vspace{0.5em}

\noindent
\textbf{B$^3$} \cite{bagga1998algorithms}. Mention-level measure which computes precision and recall for each individual mention, overcoming the singleton problem of MUC, but can problematically count the same coreference chain multiple times.
\vspace{0.5em}

\noindent
\textbf{CEAF-M} \cite{luo2005coreference}. Mention-level measure which reflects the percentage of mentions that are in the correct coreference chains. Note that precision and recall are the same in this measure since we use pre-annotated mentions.
\vspace{0.5em}

\noindent
\textbf{CEAF-E} \cite{luo2005coreference}. Entity-level measure computed by aligning predicted with the gold chains, not allowing one chain to have more than one alignment, overcoming the problem of B$^3$.
\vspace{0.5em}

\noindent
\textbf{BLANC} \cite{luo2014extension}. Computes two F-scores in terms of the pairwise quality of coreference decisions and non-coreference decisions, and averages these scores together for the final results.
\vspace{-0.75em}

\noindent
\textbf{CoNLL}. The mean of MUC, B$^3$, and CEAF-E.

\subsection{Models}
We compare our representation-learning model variants to three baselines: a deterministic lemma-based baseline, a lemma-$\delta$ baseline, and an unsupervised baseline which clusters the originally extracted features. We also compare with the results of \citeauthor{cybulska2015translating} \shortcite{cybulska2015translating}.

\subsubsection{Baselines}

\noindent
\textbf{\textsc{Lemma}}. This algorithm clusters event mentions which share the same head word lemma into the same coreference chains across all documents.
\vspace{0.5em}

\noindent
\textbf{\textsc{Lemma-$\delta$}}. Proposed by \citeauthor{upadhyay2016revisiting} \shortcite{upadhyay2016revisiting}, this method provides a difficult baseline to beat. A $\delta$-similarity threshold is introduced, and we merge two mentions with the same head-lemma if and only if the cosine-similarity between the TF-IDF vectors of their corresponding documents is greater than $\delta$. This $\delta$ parameter is tuned to maximize B$^3$ performance on the validation set, which we found occurs when $\delta=0.67$.
\vspace{0.5em}

\noindent
\textbf{\textsc{Unsupervised}.} This is the result obtained by agglomerative clustering over the original unweighted features. Again, we optimize the $\tau$ similarity threshold over the validation set.

\subsubsection{Sentence Templates (CV2015)}  \citeauthor{cybulska2015translating} \shortcite{cybulska2015translating} propose a model that uses sentence-level event templates (see Table \ref{tab:eventTemplate}), requiring more annotated information than our models. See \cite{vossen2017identity} for further elaboration of this model. To our knowledge, this is the best previous model on ECB+ using the same data and evaluation criteria as our work.


\subsubsection{Representation Learning.}
We test four different model variants:
\begin{itemize}[leftmargin=1em]
    \item \textsc{CCE}: uses only categorical-cross-entropy in the loss function (Equation~\ref{eq:cce}); 
    
    \item \textsc{CORE}: uses only clustering-oriented regularization; i.e., the attract and repulse terms (Equations~\ref{eq:attract} and \ref{eq:repulse}); 
    
    \item \textsc{CORE+CCE}: includes categorical-cross-entropy and the attract and repulse terms (Equation~\ref{eq:reg}); 

    \item \textsc{CORE+CCE+Lemma}: initializes the agglomerative clustering with clusters computed by lemma-$\delta$ (with a differently tuned value of $\delta$ than the baseline) and continues the clustering process using the similarities between the embeddings created by \textsc{CORE+CCE}.
\end{itemize}

\subsection{Hyper-parameter Tuning}

\begin{table}[t]
    \begin{center}
        \begin{tabular}{l|c|c|c|c}
        \toprule
        \textbf{Model} & $\lambda_1$ & $\lambda_2$ & B$^3$ & $\tau$\\
        \hline

        \textbf{Baselines}&&&&\\
        \textsc{Unsupervised} & - & - & $0.590$ & $0.657$ \\
        \textsc{Lemma} & - & - & $0.597$ & - \\
        \textsc{Lemma}-$\delta$ & - & - & $0.612$ & - \\
        \hline
        
        \textbf{Model Variants} &&&&\\
        
        \textsc{CORE+CCE+L} & 2.0 & 0.0 & $\textbf{0.678}$ & $0.843$ \\
        \hdashline
        
        \textsc{CORE+CCE}
         & $2.0$ & $2.0$  & $0.663$ & $0.776$ \\
         & $2.0$ & $1.0$  & $0.666$ & $0.773$\\
         & $2.0$ & $0.1$  & $0.665$ & $0.843$ \\
         & $2.0$ & $0.0$  & $\textbf{0.669}$ & $0.843$\\ 
         & $0.0$ & $2.0$  & $0.662$ & $0.710$ \\
         \hdashline
        
        \textsc{CORE}  & $2.0$ & $2.0$ & $0.631$ & $0.701$ \\
             & $1.0$ & $1.0$ & $0.625$ & $0.689$ \\
    
        \hdashline
        \textsc{CCE} & - & - & $0.644$ & $0.853$ \\ 
        
        \bottomrule
        \end{tabular}
        
    \end{center}
    \caption{Model comparison based on validation set B$^3$ accuracy with optimized $\tau$ cluster-similarity threshold. For \textsc{CORE+CCE+Lemma} (indicated as \textsc{CORE+CCE+L}) we tuned to $\delta=0.89$; for \textsc{Lemma}-$\delta$ we tuned to $\delta=0.67$.}\label{tab:valopt}
\end{table}

\begin{table*}[t]
    {\small
    \centering
    \hfill{}
    \begin{tabular}{l|ccc|ccc|c|ccc|ccc|c}
        \toprule
         & & MUC & & & B$^3$ & & CM & & CE & & & BLANC & & \textsc{CoNLL}\\
        \textbf{Model} & R & P & F & R & P & F & F & R & P & F & R & P & F & F\\
        \hline
        
        \textbf{Baselines} &&&&&&&&&&&&&&\\
        \textsc{Lemma}             & 66 & 58 & 62 & 66 & 58 & 62 & 51 & 87 & 39 & 54 & 64 & 61 & 63 & 61\\
        \textsc{Lemma-$\delta$}    & 55 & 68 & 61 & 61 & 80 & \textbf{69} & \textbf{59} & 73 & 60 & 66 & 62 & 80 & \textbf{67} & 66\\
        \textsc{Unsupervised}      & 39 & 63 & 48 & 55 & 81 & 66 & 51 & 72 & 49 & 58 & 57 & 58 & 58 & 57\\
        \hline
        
        \textbf{Previous Work} &&&&&&&&&&&&&&\\
        CV2015          & {43} & {77} & {55} & {58} & {86} & \textbf{{69}} & 58 & -  & -  & {66} & {60} & {69} & 63 & {64}\\
        \hline
        
        \textbf{Model Variants} &&&&&&&&&&&&&&\\
        \textsc{CCE} & 66 & 63 & 65 & 69 & 60 & 64 & 50 & 59 & 63 & 61 & 69 & 56 & 59 & 63\\
        \textsc{CORE} & 58 & 58 & 58 & 66 & 58 & 62 & 44 & 53 & 53 & 53 & 66 & 54 & 56 & 57 \\
        \textsc{CORE+CCE} & 62 & 70 & 66 & 67 & 69 & 68 & 56 & 73 & 64 & 68 & 68 & 59 & 62 & 67\\
        \textsc{CORE+CCE+Lemma} & 67 & 71 & \textbf{69} & 71 & 67 & \textbf{69} & 58 & 71 & 67 & \textbf{69} & 72 & 60 & 64 & \textbf{69}\\
        \bottomrule
        
    \end{tabular}}
    \hfill{}
    \caption{Combined within- and cross-document test set results on ECB+. Measures CM and CE stand for mention-based CEAF and entity-based CEAF, respectively.}
    \label{tab:testresults}
\end{table*}

\begin{table*}[t]
    {\small
    \centering
    \hfill{}
    \begin{tabular}{l|ccc|ccc|c|ccc|ccc|c}
        \toprule
         & & MUC & & & B$^3$ & & CM & & CE & & & BLANC & & \textsc{CoNLL}\\
        \textbf{Model} & R & P & F & R & P & F & F & R & P & F & R & P & F & F\\
        \hline
        
        \textbf{Baselines} &&&&&&&&&&&&&&\\
        \textsc{Lemma-$\delta$}    & 41 & 77 & 53 & 86 & 97 & \textbf{92} & 85 & 92 & 82 & 87 & 65 & 86 & 71 & 77\\
        \textsc{Unsupervised}      & 32 & 36 & 34 & 85 & 86 & 85 & 74 & 80 & 78 & 79 & 65 & 55 & 57 & 66\\
        
        \hline
        
        \textbf{Model Variants} &&&&&&&&&&&&&&\\
        \textsc{CCE} & 44 & 49 & 46 & 87 & 89 & 88 & 79 & 82 & 80 & 81 & 67 & 67 & 67 & 72\\
        \textsc{CORE} & 55 & 32 & 40 & 89 & 70 & 78 & 65 & 64 & 79 & 71 & 75 & 54 & 56 & 63\\
        \textsc{CORE+CCE} & 43 & 68 & 53 & 87 & 95 & 91 & 84 & 90 & 82 & 86 & 67 & 76 & 70 & 76\\
        \textsc{CORE+CCE+Lemma}          & 57 & 69 & \textbf{63} & 90 & 94 & \textbf{92} & \textbf{86} & 90 & 86 & \textbf{88} & 73 & 78 & \textbf{75} & \textbf{81}\\
        \bottomrule
    \end{tabular}}
    \hfill{}
    \caption{Within-document test set results on ECB+. Note that \textsc{Lemma} is equivalent to \textsc{Lemma-$\delta$} in the within-document setting. \citeauthor{cybulska2015translating} \shortcite{cybulska2015translating} did not report the performance of their model in this setting.}
    \label{tab:withinresults}
\end{table*}

\noindent
For the representation learning models, we performed a non-exhaustive hyper-parameter search optimized for validation set performance. We keep the following parameters constant across the model variants:


\begin{itemize}[leftmargin=1em,]
    \item $1000$ neurons in H$_1$ and H$_3$; $250$ neurons in H$_e$, the embedding layer (see Figure \ref{fig:train_model});
    \item Softmax output layer with $C+1$ units;
    \item ReLU activation functions for all neurons;
    \item \textit{Adam} gradient descent \cite{kingma2014adam};
    \item $25\%$ dropout between each layer;
    \item Learning rate of $0.00085$ (times $10^{-1}$ for \textsc{CORE});
    \item Randomly sampled batches of $272$ mentions, where a batch is forced to contain pairs of coreferential and non-coreferential mentions.
\end{itemize}

Models are trained for 100 epochs. At each epoch, we optimize $\tau$ (our agglomerative clustering similarity threshold) using a two-pass approach: we first test $20$ different settings of $\tau$, then $\tau$ is further optimized around the best value from the first pass. For \textsc{CORE+CCE+Lemma}, we tune the $\delta$ parameter of the lemma-$\delta$ clustering approach to the validation set by testing $100$ different values of $\delta$; these different $\delta$ values initialize the clusters, and we then continue clustering by testing validation results obtained when using the similarities between the embeddings created by \textsc{CORE+CCE} for different values of $\tau$.

Some of the results of hyperparameter tuning on the validation set are shown in Table~\ref{tab:valopt}. Interestingly, we observe that \textsc{CORE+CCE} performs slightly better with $\lambda_2=0$; i.e., without repulsive regularization. This suggests that enforcing representation similarity is more important than enforcing division, although we cannot conclusively state that repulsive regularization would not be useful for other tasks. Nonetheless, for test set results we use the optimal hyperparameter configurations found during this validation-tuning step; e.g., for \textsc{CORE+CCE} we set $\lambda_1=2$ and $\lambda_2=0$.

\section{Results}

Table~\ref{tab:testresults} presents the performance of the models for combined within- and cross-document event coreference. Results for these models are obtained with the hyper-parameter settings that achieved optimal accuracy during validation-tuning.

Firstly, we observe that \textsc{CORE+CCE} offers marked improvements upon the \textsc{Unsupervised} baseline, \textsc{CORE} model, and \textsc{CCE} model. From these results we conclude: (i) supervised representation learning provides more informative embeddings than the original feature vectors; and, (ii) that combining Clustering-Oriented Regularization with categorical-cross-entropy is better than just using one or the other, indicating that our introduction of these novel terms into the loss function is a useful contribution.

We also note that \textsc{CORE+CCE+Lemma} (which obtains the best validation set results) beats the strong \textsc{Lemma}-$\delta$ baseline. Our model offers marked improvements or roughly equivalent scores in each evaluation measure except BLANC, where the baseline offers a $3$ point F-score improvement. This is due to the very high precision of the baseline, whereas \textsc{CORE+CCE+Lemma} seems to trade precision for recall.

We finally observe that \textsc{CORE+CCE+Lemma} improves upon the results of \citeauthor{cybulska2015translating} \shortcite{cybulska2015translating}. We obtain improvements of $14$ points in MUC, $3$ points in entity-based CEAF, $5$ points in CoNLL, and $1$ point in BLANC, with equivalent results in B$^3$ and mention-based CEAF. These results suggest that high quality coreference chains can be built without necessitating event templates.

In Table~\ref{tab:withinresults}, we see the performance of our models on within-document coreference resolution in isolation. These results are obtained by cutting all links drawn across documents for the gold standard chains and the predicted chains. We observe that, across all models, scores on the mention- and entity-based measures are substantially higher than the link-based measures (e.g., MUC and BLANC). The usefulness of \textsc{CORE+CCE+Lemma} (which initializes the clustering with the lemma-$\delta$ predictions and then continues to cluster with \textsc{CORE+CCE}) is exemplified by the improvements or matches in every measure when compared to both \textsc{Lemma-$\delta$} and \textsc{CORE+CCE}. The most vivid improvement here is observed with the $10$ point improvement in MUC over both models as well as the $4$ and $5$ point improvements in BLANC respectively, where the higher recall entails that \textsc{CORE+CCE+Lemma} confidently predicts coreference links that would otherwise have been false negatives.

\section{Conclusions and Future Work}
We have presented a novel approach to event coreference resolution by combining supervised representation learning with non-parametric clustering. We train an hourglass-shaped neural network to learn how to represent event mentions in a useful way for an agglomerative clustering algorithm. By adding the novel Clustering-Oriented Regularization (CORE) terms into the loss function, the model learns to construct embeddings that are easily clusterable; i.e., the prior that embeddings of samples belonging to the same cluster should be similar, and those of samples belonging to different clusters should be dissimilar.


Our results suggest that clustering embeddings created with representation learning is much better than clustering of the original feature vectors, when using the same agglomerative clustering algorithm. We show that including CORE in the loss function improves performance more than when only using categorical-cross-entropy to train the representation learner model. Our top-performing model obtains results that improve upon previous work despite the fact that our model requires less annotated information in order to perform the task. 

Future work involves applying our model to automatically annotated event mentions and other event coreference datasets, and extending this framework toward a full end-to-end system that does not rely on manual feature engineering at the input level. Additionally, our model may be useful for other clustering tasks, such as entity coreference and document clustering. Lastly, we seek to determine how CORE and its imposition of a clusterable latent space structure may or may not assist in improving the quality of latent representations in general.

\section*{Acknowledgements}
This work was funded with grants from the Natural Sciences and Engineering Research Council of Canada (NSERC) and the Fonds de recherche du Qu\'ebec –- Nature et Technologies (FRQNT). We thank the anonymous reviewers for their helpful comments and suggestions.

\bibliographystyle{acl_natbib}
\bibliography{naaclhlt2018.bib}

\end{document}